\documentclass[runningheads]{llncs}
\usepackage{cite}
\usepackage{amsmath,amssymb,amsfonts}
\usepackage{algorithmic}
\usepackage{graphicx}
\usepackage{textcomp}
\usepackage{xcolor}
\usepackage{url}
\usepackage{booktabs}
\usepackage{multirow}
\usepackage{threeparttable}
\usepackage[hidelinks]{hyperref}
\usepackage[utf8]{inputenc}
\usepackage[T1]{fontenc}
\usepackage{graphicx}
\begin{document}

\title{\large Reliable Academic Conference Question Answering: \\A Study Based on Large Language Model}
\titlerunning{Reliable Academic Conference QA}
%
\author{Zhiwei Huang\inst{1} \and Juan Li\inst{1} \and Long Jin\inst{1} \and  Junjie Wang\inst{1} \and Mingchen Tu\inst{1} \and Yin Hua\inst{1} \and Zhiqiang Liu\inst{1} \and Jiawei Meng\inst{2}\and Wen Zhang\inst{1}\thanks{Corresponding author.}}
\authorrunning{Z. Huang et al.}
%
\institute{School of Software, Zhejiang University \and
College of Computer Science and Technology, Zhejiang University \\
\email{\{huangzww, lijuan18, longjin, wangjj2018, mingchentz, 22351088, zhiqiangliu, mjw.cs, zhang.wen\}@zju.edu.cn}
}

\maketitle              
\begin{abstract}
\begin{sloppypar}
As the development of academic conferences fosters global scholarly communication, 
researchers consistently need to obtain accurate and up-to-date information about academic conferences. 
Since the information is scattered, using an intelligent question-answering system to efficiently handle researchers' queries and ensure awareness of the latest advancements is necessary. Recently, Large Language Models (LLMs) have demonstrated impressive capabilities in question answering, and have been enhanced by retrieving external knowledge to deal with outdated knowledge. 
However, these methods fail to work due to the lack of the latest conference knowledge. To address this challenge, we develop the ConferenceQA dataset, consisting of seven diverse academic conferences. Specifically, for each conference, we first organize academic conference data in a tree-structured format through a semi-automated method. 
Then we annotate question-answer pairs and classify the pairs into four different types to better distinguish their difficulty.
With the constructed dataset, we further propose a novel method STAR (\textbf{ST}ructure-\textbf{A}ware \textbf{R}etrieval) to improve the question-answering abilities of LLMs, leveraging inherent structural information during the retrieval process. 
Experimental results on the ConferenceQA dataset show the effectiveness of our retrieval method.
The dataset and code are available at https://github.com/zjukg/ConferenceQA.
\end{sloppypar}
\end{abstract}
\vspace{-7mm}
\section{Introduction}
The rapid advancement of computer science has led to an increase in research presented at academic conferences, which are crucial for academic exchange. Given the vast and dispersed nature of conference information, querying is a more efficient method for information retrieval than navigating multiple sources.

\par
Recent advancements in Large Language Models (LLMs) \cite{brown2020language,openai2023gpt4,touvron2023llama2} have significantly impacted various NLP tasks, including question answering. LLMs demonstrate capabilities like chain-of-thought reasoning\cite{wei2022chain} and in-context learning\cite{min2022rethinking}, enhanced by increasing model parameters and extensive training data. After instruction fine-tuning \cite{chung2022scaling}, LLMs excel in conversational tasks and information retrieval\cite{kojima2022large}.
\par 

Despite the success of LLMs, they are related to incompleteness, untimeliness, unfaithfulness, and having limitations in updating timely and domain-specific expertise. 
This necessitates research efforts to integrate LLMs with external knowledge sources, such as knowledge bases (KBs)\cite{modarressi2023retllm}, search engines\cite{schick2023toolformer} and databases\cite{hu2023chatdb}. 
Regarding academic conference queries, due to the missing external conference knowledge, LLMs fail to access the latest academic conference information in question answering, such as academic conferences in 2022 and later ones. 
Existing retrieval methods are efficient but primarily focus on plain text\cite{izacard2022atlas}, triples\cite{sen2023knowledge}, and tables \cite{zhong2017seq2sql}, which does not align well with the structured nature of conference websites, complicating direct application for conference-specific queries.
\par
In this paper, we introduce ConferenceQA, a benchmark comprising seven recent top-tier academic conferences, these conferences span various research domains such as web science, natural language processing, machine learning, databases, artificial intelligence, and the semantic web, providing a comprehensive dataset that organizes information across all stages of the conferences.
To construct this dataset, we initially employ a semi-automatic method to convert the conference information into a tree structure. Subsequently, we utilize ChatGPT to simulate roles with diverse backgrounds, enabling us to generate role-specific questions. These questions are then carefully filtered and annotated with answers to ensure the dataset's reliability. Additionally, we document the sources of the answers to further enhance the dataset's credibility.
Besides, we categorize the questions into four types given the complexity of getting answers.

On the constructed ConferenceQA dataset, we introduce STAR (\textbf{ST}ructure-\textbf{A}ware \textbf{R}etrieval), a method leveraging LLMs for hierarchical data, and then proceed to conduct a study on conference QA.
Our method generates a textual description for each path based on both its surrounding structural information and its own textual information. 
We conduct experiments using various LLMs, 
along with different retrievers. 
Compared to path retrieval, structural-aware retrieval shows an average relative F1 score improvement of 15.50\% across different LLMs and 17.03\% when using different retrievers. This highlights the effectiveness of STAR on the tree-structured ConferenceQA dataset. 
\begin{figure*}[!htb]
\centering
\includegraphics[width=0.8\linewidth]{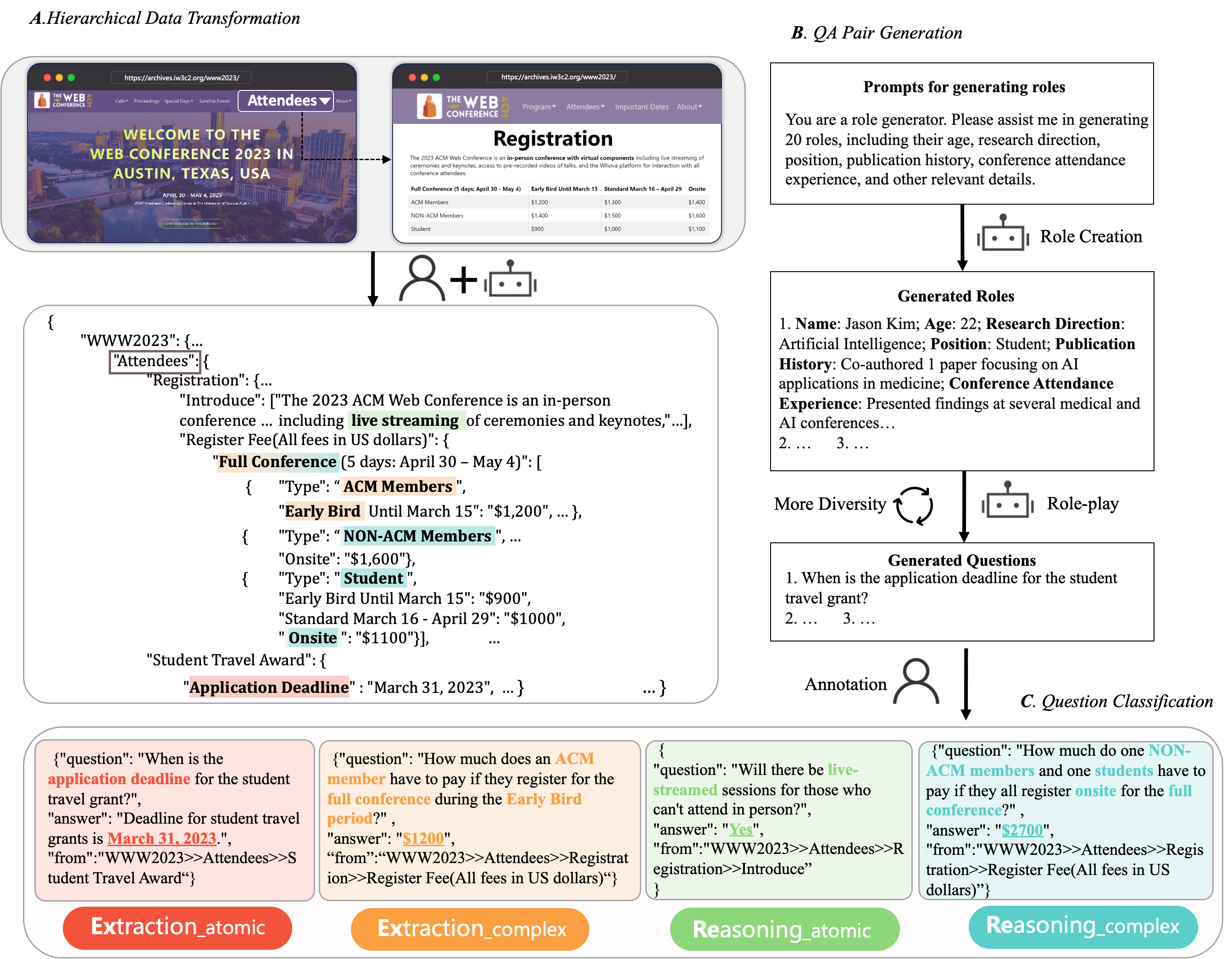}
\caption{The illustration of the ConferenceQA dataset construction process. Initially, data from official websites is transformed into a tree structure semi-automatically. Next, questions are generated through role-play and manually annotated with answers. Finally, questions are categorized by the complexity of paths and reasoning required.
}
\label{fig:dataset-construction}
\end{figure*}
Our contributions can be summarized as follows:
\vspace{-1mm}
\begin{enumerate}
    \item {We construct a benchmark called ConferenceQA, organizing conference information in a tree structure, to assist evaluate question answering about academic conferences.}
    \item {We introduce a novel method called STAR. By utilizing the structural information around nodes to generate textual descriptions, and using these descriptions for retrieval, it can effectively enhance answer performance.}
    \item {
    We conduct experiments on the ConferenceQA dataset,  proving that LLMs enhanced with retrieval methods could successfully answer questions about academic conferences and our STAR method consistently outperforms path retrieval method, offering meaningful insights.
    
    } 
\end{enumerate}
\vspace{-10mm}
\section{Dataset Construction}
In this section, we introduce the construction of the ConferenceQA dataset. We select the conference information of seven typical academic conferences in 2022 or 2023 to build the dataset based on their official website, where the most accurate information about the conferences is stored. Each conference is assigned to one data annotator with relevant experience in the realm of academic conferences.
We use three steps, including \textit{hierarchical data transformation}, \textit{QA pair generation} and \textit{question classification}, to construct each conference dataset. The overview of the construction process is shown in Fig.~\ref{fig:dataset-construction}.
\vspace{-4mm}
\subsection{Hierarchical Data Transformation} 
Data transformation in the ConferenceQA dataset involves standardizing the diverse formats of academic conference data sourced from official conference websites into a unified tree structure. Each conference page combines unstructured text, like conference introductions and paper submission guidelines, with structured data such as payment and schedule details. To manage this format variability, we employ a semi-automated method to create tree-structured data for each conference.

Specifically, the automated component converts structured table data into a tree format using ChatGPT, as shown in Fig.~\ref{fig:dataset-construction}, where registry information is transformed. For other structured data, such as accepted papers with consistent schemas (title, authors, abstract), we employ web crawlers to fetch HTML pages and convert them into corresponding tree-structured data based on the HTML tags. The manual component involves annotating inter-page relationships. Annotators assign page titles to tree nodes based on the linkage among pages, evident in navigation bars and subpage links like `calls', `proceedings' and `programs'. Additionally, subtitles within pages are identified and designated as child nodes under the relevant page titles. These manual steps are essential to maintain the dataset's quality and coherence.

Ultimately, we obtain seven conference datasets organized in a tree-structured format. They are served as accurate and rigorous knowledge sources. 
\vspace{-2mm}
\subsection{QA Pair Generation} This step involves generating reliable question-answer pairs through role creation, LLM-generated questions, and manual annotation. For each conference, we utilize ChatGPT to simulate the roles of conference participants, generating relevant questions which are then manually filtered and annotated with answers and their sources to ensure realism and reliability.

We use ChatGPT to create 20 roles characterized by specific attributes such as age, research direction, position, publication history, and conference attendance experience, mimicking real-life researchers with diverse backgrounds interested in the conferences.
With these roles, we prompt ChatGPT to engage in role-playing scenarios, generating five varied questions per conference. These questions cover different areas of interest or uncertainty relevant to the roles' diverse backgrounds. To avoid redundancy and enhance question diversity, we iteratively prompt the model. Specifically, we use the results generated by the ChatGPT as examples for the next iteration and encourage the ChatGPT to generate more diverse questions. In the final step, we manually review and filter the questions to eliminate duplicates and unrealistic queries. We then annotate the answers based on our tree-structured data, ensuring the reliability of the dataset by documenting the source of each answer within the constructed academic conference data.
\vspace{-2mm}
\subsection{Question Classification} To assess the model's capability in handling questions of varying difficulty, we design a scheme to classify the question-answer pairs based on two criteria: the method used to generate the answer and the complexity of paths required to arrive at the correct answer.

\textbf{Extraction vs. Reasoning:} This category evaluates the process of answer generation. Answers directly pulled from the dataset are labeled as \textit{extraction}, whereas answers that necessitate reasoning beyond the dataset content are labeled as \textit{reasoning}. \textit{Reasoning} questions are more challenging than \textit{extraction} questions because, unlike direct extraction, \textit{reasoning} questions require the model to have the capability to infer the relationship between the retrieved paths and the question.

\textbf{Atomic vs. Complex:} This category assesses the complexity of paths needed to generate the answer. Answers that depend on a single path are termed \textit{atomic}, while those requiring multiple paths are termed \textit{complex}. \textit{Complex} questions are more difficult than \textit{atomic} questions because, instead of a single path, \textit{complex} questions require recalling multiple paths to derive an answer. 

 Combining these dimensions results in four levels of difficulty: \textit{extraction-atomic}, \textit{extraction-complex}, \textit{reasoning-atomic}, and \textit{reasoning-complex}. This classification is vital for analyzing the model's performance across different complexities and reasoning demands.

\vspace{-2mm}
 \subsection{Dataset Validation}
Following data construction, a thorough validation process is conducted by three independent assessors who evaluate each QA pair across three critical dimensions. The first dimension assesses the alignment between each question and its answer, ensuring the answer accurately addresses the question. Concurrently, the second dimension examines the reliability of the answer source, ensuring it provides the necessary information for the question. The third dimension evaluates the practical relevance of each question, ensuring it reflects real-world needs and concerns. If a QA pair fails to meet the criteria in any dimension, as agreed upon by at least two assessors, it is marked for removal and redesign. This rigorous process ensures each QA pair is validated comprehensively, maintaining the quality and reliability of the dataset. Detailed statistics of the selection process for each conference are shown in Table \ref{tab:dataset}.
\vspace{-4mm}
\begin{center}
\begin{table}[ht]
\centering
\caption{Statistics of the ConferenceQA dataset. \#Paths indicates the number of tree branches, and \#Depth shows the depth of the tree. \#EA, \#EC, \#RA, and \#RC represent extraction-atomic, extraction-complex, reasoning-atomic, and reasoning-complex question types, respectively.}
\begin{tabular}{ccccccc}
\hline
Conference & \#Paths & \#Depth & \#EA & \#EC & \#RA & \#RC\\ \hline
\href{https://archives.iw3c2.org/www2023/}{WWW2023} & 15127 & 7.01 & 32 & 27 & 17 & 36  \\
\href{https://2023.aclweb.org/}{ACL2023} & 14306 & 9.05 & 29& 21& 30& 25 \\
\href{https://icml.cc/Conferences/2023}{ICML2023} & 4715  & 8.52 & 26& 27& 28& 19 \\
\href{https://2023.sigmod.org/}{SIGMOD2023} & 6338 & 7.46 & 39& 27& 23& 34\\
\href{https://ijcai-23.org/#}{IJCAI2023} & 15800  & 6.13 & 28& 26& 13&33  \\
\href{https://icde2023.ics.uci.edu/}{ICDE2023} & 9736  & 9.14 & 28&24 &22 &21  \\
\href{https://iswc2022.semanticweb.org/}{ISWC2022} & 3594  & 7.53 & 33& 42& 25& 18 \\ \hline
Avg & 9916 & 7.83 & 31 & 28 &23 & 27\\ \hline
\label{tab:dataset}
\end{tabular}
\end{table}
\end{center}
\vspace{-18mm}
\section{Method}

In this section, we discuss LLM-based methods for academic conference question-answering. The prevalent approach involves using an external knowledge source for retrieval\cite{sen2023knowledge, hu2023chatdb, shi2023replug}, where the reader's query $q$ extracts relevant content $c$ from a domain-specific knowledge base, and this content is then combined with the query for the LLM to generate an answer. This retrieval-based method can be formalized as $a = LLM(q, c)$ where $c = Retriever(q, \mathcal{KB})$. It optimizes the retriever, such
that for each question $q$, the model can give an answer $a$ that has high accuracy or relevancy with a correct answer.
Our approach adheres to this retrieval-based model but is adapted for our conference's tree-structured dataset. We preprocess this structured data to facilitate content retrieval and introduce a novel method named STAR (\textbf{ST}ructure-\textbf{A}ware \textbf{R}etrieval), which effectively integrates structural and semantic data for improved retrieval performance.
\vspace{-2mm}
\subsection{Tree-structured Data Processing}
The tree-structured data is hierarchically arranged, with each node representing a page or a section heading, and each leaf node corresponding to its specific content. For retrieval, we pair each leaf node with its root node to provide additional context to the LLM. Paths in the tree use the `>>' field to denote hierarchical relationships and contain both structural and semantic information. An example path is: \textit{WWW2023>>Attendees>>Registration>>Register Fee>>Virtual Conference>>ACM Members>>\$300}. 
After the tree-structured data processing, the knowledge source for retrieval could be represented as a set of paths that $\mathcal{P} = \{p_1, p_2, ..., p_m \}$ where $m$ is the number of paths in the dataset.
\vspace{-2mm}
 \subsection{Path Retrieval}
Upon receiving a query input $q$, the retriever selects a subset of paths from $\mathcal{P} = \{p_1, p_2, ..., p_m \}$ that are relevant to $q$. 
Following established methods \cite{ni2021large}, we use a dense retriever based on a dual encoder framework. This framework employs an encoder to transform both the query $q$ and each path 
$p \in \mathcal{P}$ into embeddings. The similarity between the query and path embeddings is assessed using cosine similarity, and the top-$k$ paths with the highest similarity scores are retrieved, as expressed in \eqref{retrieve}, where \textbf{E} denotes the embedding function.

\begin{equation}
    c = topk(\{ \cos(\textbf{E}(q), \textbf{E}(p)) | p \in \mathcal{KP}\})
\label{retrieve}
\end{equation} 

\begin{figure*}[htbp]
\includegraphics[width=0.8\linewidth]{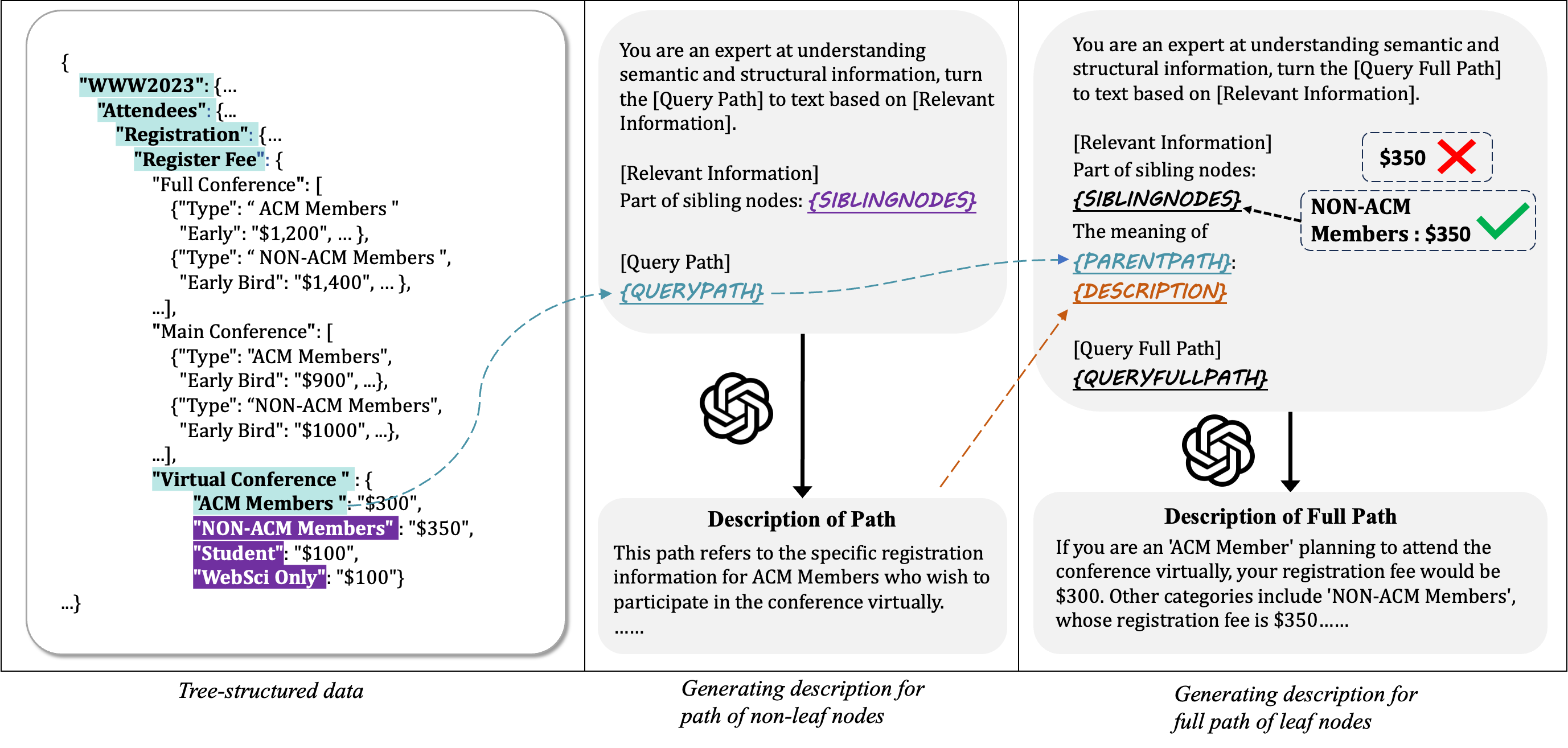}
\centering
\caption{The diagram depicts the iterative, top-to-bottom generation of path descriptions using ChatGPT, which incorporates surrounding structural information.}
\label{fig:desp_of_path}
\end{figure*}
\vspace{-2mm}
\subsection{Structure-aware Retrieval}
The limitation of treating a single path as the retrieval object is that it disconnects the structural relationships among paths. For example, the relationship between an author's name and their affiliated institution is lost when paths are retrieved independently.

To overcome this, we introduce a novel method called STAR (\textbf{ST}ructure-\textbf{A}ware \textbf{R}etrieval). As shown in Fig.~\ref{fig:desp_of_path}, STAR employs ChatGPT to iteratively generate textual descriptions for each path $des_p$, from the root to individual nodes, in a top-down manner. We enhance the retrieval process by incorporating structural information in the user input, which includes siblings, parent path descriptions, and the query path itself. This approach helps maintain the contextual relevance of each path, which is crucial for recognizing relationships like those between an author and their institution. For instance, when generating path descriptions, we not only consider the node's immediate context but also integrate the structural significance of related nodes. This includes the siblings of a node and their parent nodes, ensuring a comprehensive representation of each path's context. To avoid the loss of information about the siblings of leaf nodes, we append the text of their parent node to each sibling of the leaf nodes. Ultimately, this method effectively preserves and utilizes structural relationships, enhancing the retrieval process.

Thus we can construct a knowledge source of path descriptions $\mathcal{KP}_{des_p} = \{ (p, des_p) | p \in \mathcal{KP}\}$, containing pairs of paths and their descriptions. For retrieval, we use the similarity between the query and each path description as the score for that path. We then retrieve the top-$k$ paths with the highest similarity scores to the query $q$. The embedding of the element is denoted by \textbf{E}, and this process is formalized as shown in \eqref{redesc}.
\begin{equation}
    c = topk(\{ \cos(\textbf{E}(q), \textbf{E}(des_p)) | (p, des_p) \in \mathcal{KP}_{des_p}\})
\label{redesc}
\end{equation}

\vspace{-4mm}
\section{Experiments}
In this section, we conduct question answering experiments on conference datasets to explore: 1) How does the STAR perform with different LLMs? 2) How does the STAR perform with different retrievers? 3) How does the STAR perform with different academic conferences?
\vspace{-2mm}
\subsection{ Experimental Details}

Based on the constructed ConferenceQA, we use currently popular LLMs, including Bloom (7B) \cite{le2022bloom}, GPT-J (6B) \cite{wang2021gpt}, Flan-T5 (xl and xxl) \cite{longpre2023flan}, LLaMA2 (7B and 13B) \cite{touvron2023llama2}, Mistral (7B) \cite{jiang2023mistral} and ChatGPT, as the main evaluation backbone to assess the performance of mainstream LLMs. For ChatGPT, we employ GPT-3.5-turbo and access it via API\footnote{ from https://api.openai.com/}. 
We employ BM25\cite{robertson2009probabilistic}, SentenceBert\cite{reimers2019sentence}, DPR\cite{karpukhin2020dense}, ANCE\cite{xiong2020approximate} and text-embedding-ada-002 as our retriever.
In addition, we use Chroma\footnote{https://github.com/chroma-core/chroma} as our vector database and employ cosine similarity for matching. In all experiments, we select the top 5 paths retrieved. 

\vspace{-2mm}
\subsection{Evaluation Metrics} In line with prior studies, we assess the QA capabilities of LLMs using the F1 score and the exact match (EM) score. Specifically, we employ GPT-4 to compute the EM, referred to as EM-GPT4.

The F1 score quantifies the overlap between the predicted and correct answers by calculating the harmonic mean of precision and recall.

The EM-GPT4 score evaluates the proportion of instances where the LLM's predicted answer exactly matches the correct answer. Given the generative nature of LLMs, slight textual variations in responses might still represent the same answer. We use GPT-4, a highly advanced LLM known for its semantic understanding capabilities, to precisely assess if the LLM's response matches the golden answers. 

\begin{table*}[t]
\centering
\renewcommand{\arraystretch}{1.5} 
\caption{EM-GPT4 and F1 scores for various LLMs in the ConferenceQA dataset, categorized by question types: EA (extraction-atomic), EC (extraction-complex), RA (reasoning-atomic), and RC (reasoning-complex). The retriever used is text-embedding-ada-002. Black numbers show path retrieval performance. ‘+’ denotes performance improvement with the STAR method.
\textcolor{red!70!black}{Red} indicates positive enhancements, \textcolor{green!70!black}{Green} signifies reductions.}


\label{tab:results}
\resizebox{\textwidth}{!}
{
\begin{tabular}{c|cccc|cccc}
\cline{1-9}
\multirow{2}{*}{LLMs} & \multicolumn{4}{c|}{F1} & \multicolumn{4}{c}{EM-GPT4}\\ \cline{2-9}
 & EA & EC & RA & RC & EA & EC & RA  & RC \\ \cline{1-9}
 Bloom-7B1 &19.60\textcolor{green!70!black}{-0.36} & 11.19\textcolor{red!70!black}{+1.96} & 17.01\textcolor{red!70!black}{+0.36} & 11.58\textcolor{green!70!black}{-0.20} &30.27\textcolor{red!70!black}{+0.42} & 15.03\textcolor{red!70!black}{+4.41}& 41.70\textcolor{green!70!black}{-3.43} & 17.58\textcolor{green!70!black}{-1.36} \\ \cline{1-9}
 GPT-J-6B &14.53\textcolor{red!70!black}{+1.76}& 8.81\textcolor{red!70!black}{+3.11}& 15.52\textcolor{red!70!black}{+2.46} & 8.42\textcolor{green!70!black}{-0.25} & 19.11\textcolor{red!70!black}{+7.04} & 12.93\textcolor{red!70!black}{+5.53} & 34.16\textcolor{red!70!black}{+5.03} & 13.08\textcolor{red!70!black}{+1.94}\\ \cline{1-9}
 Flan-T5-xl & 27.74\textcolor{red!70!black}{+7.85} & 14.68\textcolor{red!70!black}{+2.77} & 36.03\textcolor{red!70!black}{+0.96} & 19.00\textcolor{red!70!black}{+2.89} & 35.50\textcolor{red!70!black}{+9.56} & 20.78\textcolor{red!70!black}{+0.86} & 59.38\textcolor{red!70!black}{+3.97} & 25.74\textcolor{red!70!black}{+2.27} \\ 
 Flan-T5-xxl &32.31\textcolor{red!70!black}{+8.97} & 14.01\textcolor{red!70!black}{+9.69} & 37.08\textcolor{red!70!black}{+4.3} & 20.86\textcolor{red!70!black}{+0.98} & 40.81\textcolor{red!70!black}{+10.69} & 18.58\textcolor{red!70!black}{+12.64} & 55.76\textcolor{red!70!black}{+11.71} & 25.36\textcolor{red!70!black}{+3.58} \\ \cline{1-9}
 LLaMA2-7B & 14.05\textcolor{red!70!black}{+2.23} & 12.09\textcolor{red!70!black}{+1.25}& 12.47\textcolor{red!70!black}{+3.00} & 8.48\textcolor{red!70!black}{+0.12} & 21.32\textcolor{red!70!black}{+0.15} & 9.22\textcolor{red!70!black}{+2.77}& 23.81\textcolor{red!70!black}{+5.83} & 9.89\textcolor{green!70!black}{-1.15} \\ 
 LLaMA2-13B & 29.57\textcolor{red!70!black}{+2.82} & 20.92\textcolor{red!70!black}{+4.00} & 25.71\textcolor{red!70!black}{+4.20} & 13.64\textcolor{red!70!black}{+2.83} & 41.16\textcolor{red!70!black}{+6.26} & 24.02\textcolor{red!70!black}{+2.21} & 55.00\textcolor{red!70!black}{+6.53} & 20.23\textcolor{red!70!black}{+4.46} \\ \cline{1-9}
 Mistral-7B & 30.75\textcolor{red!70!black}{+4.31} & 23.67\textcolor{red!70!black}{+4.69} & 25.87\textcolor{red!70!black}{+4.11} & 15.91\textcolor{green!70!black}{-0.37} & 43.33\textcolor{red!70!black}{+10.53} & 27.58\textcolor{red!70!black}{+13.95} & 59.90\textcolor{red!70!black}{+6.89} & 29.23\textcolor{red!70!black}{+1.55}	\\ \cline{1-9}
 GPT-3.5-turbo & 28.35\textcolor{red!70!black}{+7.5}& 21.54\textcolor{red!70!black}{+4.83} & 24.66\textcolor{red!70!black}{+9.62} & 16.21\textcolor{red!70!black}{+0.78} & 40.53\textcolor{red!70!black}{+13.43} & 25.10\textcolor{red!70!black}{+9.03} & 49.97\textcolor{red!70!black}{+11.34} & 25.75\textcolor{red!70!black}{+1.45} \\ \cline{1-9}
 \end{tabular} 
}
\end{table*}     

\vspace{-2mm}
\subsection{Experimental Results Analysis}

\textbf{Effect of Different LLMs} We analyzed the performance of various LLMs on different types of questions to understand their perception capabilities and limitations. The results, shown in Table \ref{tab:results}, provide several insights: (1) Our STAR method significantly improves the answering performance across various LLMs. For instance, on models like Bloom-7B1, GPT-J-6B, and GPT-3.5-turbo, F1 scores increased by 4\%, 14.9\%, and 25.04\% respectively, while EM-GPT4 scores improved by 0.04\%, 24.65\%, and 24.94\%. The least improvement was on Bloom-7B1, suggesting its inherent limitations. However, substantial gains on other models demonstrate our method's effectiveness. (2) There is an inconsistency between F1 and EM-GPT4 scores; lower F1 scores sometimes align with higher EM-GPT4 scores. This may be due to LLMs generating longer textual responses, affecting F1 accuracy but not EM-GPT4, which better evaluates semantic similarity. (3) The complexity of question types affects performance; atomic questions are simpler than complex ones. Atomic questions, akin to single-hop queries, generally show higher accuracy than multi-hop complex questions. Despite this, LLMs perform comparably or better on reasoning questions than on extraction, likely due to their robust contextual learning and reasoning capabilities. (4) Different LLMs show varied understanding of paths. For example, under the same retrieval conditions, Mistral-7B outperforms GPT-3.5-turbo. Generally, models with more parameters, like LLama2-13B and Flan-T5-xxl, achieve higher accuracy, supporting the notion that larger LLMs perform better.

\begin{table*}[ht]
\centering
\renewcommand{\arraystretch}{1.5} 
\caption{EM-GPT4 and F1 scores for different retrievers in the ConferenceQA dataset. 
The generator is GPT-3.5-turbo. 
}
\begin{threeparttable}
\label{tab:retrievers_results}
\resizebox{\textwidth}{!}
{
\begin{tabular}{c|cccc|cccc}
\cline{1-9}
\multirow{2}{*}{Retrievers} & \multicolumn{4}{c|}{F1} & \multicolumn{4}{c}{EM-GPT4}\\ \cline{2-9}
 & EA & EC & RA & RC & EA & EC & RA  & RC \\ \cline{1-9}
 BM25 & 21.02\textcolor{red!70!black}{+9.14} & 16.81\textcolor{red!70!black}{+10.29} & 25.77\textcolor{green!70!black}{-2.79} & 14.72\textcolor{red!70!black}{+1.37} &25.90\textcolor{red!70!black}{+4.9}& 14.12\textcolor{red!70!black}{+8.1} & 36.50\textcolor{red!70!black}{+2.94} & 5.16\textcolor{red!70!black}{+5.55} \\ \cline{1-9}
 SentenceBERT & 38.62\textcolor{green!70!black}{-3.73} & 23.70\textcolor{green!70!black}{-0.01} & 12.97\textcolor{red!70!black}{+2.05} & 16.23\textcolor{green!70!black}{-0.18} & 39.83\textcolor{green!70!black}{-2.57} & 14.81\textcolor{red!70!black}{+11.12} & 29.79\textcolor{red!70!black}{+17.66} & 22.88\textcolor{green!70!black}{-1.47}  \\ \cline{1-9}
 DPR &  30.56\textcolor{red!70!black}{+3.72} & 23.72\textcolor{red!70!black}{+0.59} & 27.35\textcolor{red!70!black}{+2.60} & 21.16\textcolor{green!70!black}{-0.43} & 30.95\textcolor{red!70!black}{+5.91} & 15.17\textcolor{green!70!black}{-0.14} & 52.70\textcolor{red!70!black}{+0.31} & 10.66\textcolor{red!70!black}{+1.38} \\ \cline{1-9}
 ANCE & 28.24\textcolor{red!70!black}{+8.72} & 17.41\textcolor{red!70!black}{+4.75} & 16.52\textcolor{red!70!black}{+9.1} & 18.00\textcolor{red!70!black}{+2.26} & 41.66\textcolor{red!70!black}{+6.14} & 30.12\textcolor{red!70!black}{+4.07} & 50.84\textcolor{red!70!black}{+0.49} & 15.25\textcolor{red!70!black}{+1.21} \\ \cline{1-9}
 ada-002 & 28.35\textcolor{red!70!black}{+7.5}& 21.54\textcolor{red!70!black}{+4.83} & 24.66\textcolor{red!70!black}{+9.62} & 16.21\textcolor{red!70!black}{+0.78} & 40.53\textcolor{red!70!black}{+13.43} & 25.10\textcolor{red!70!black}{+9.03} & 49.97\textcolor{red!70!black}{+11.34} & 25.75\textcolor{red!70!black}{+1.45} \\ \cline{1-9}
 
 \end{tabular} 
}
    \end{threeparttable}
\end{table*}

\textbf{Effect of Different Retrievers} We evaluated four retrievers—BM25 \cite{robertson2009probabilistic}, SentenceBert \cite{reimers2019sentence}, DPR \cite{karpukhin2020dense}, and ANCE \cite{xiong2020approximate}—using the gpt-3.5-turbo generator across four question types within the ConferenceQA dataset. The results, detailed in Table \ref{tab:retrievers_results}, reveal: (1) BM25 showed weak performance, especially with \textit{extraction-atomic} and \textit{reasoning-complex} questions. In contrast, dense retrievers like SentenceBERT, DPR, and ANCE significantly outperformed BM25, underscoring the advantages of dense retrieval methods. (2) Performance varied among dense retrievers: SentenceBERT was effective in \textit{extraction-atomic} questions but less so in \textit{reasoning-atomic} questions. DPR excelled in \textit{reasoning-atomic} questions, while ANCE showed consistent performance across all question types. This indicates that selecting an appropriate retriever can significantly impact question-answering effectiveness. (3) While STAR occasionally had negative effects in some configurations, it generally enhanced performance across most settings, demonstrating its utility and reliability.


\begin{figure}
  \centering
  \label{fig:conferences}
  \includegraphics[width=0.7\linewidth]{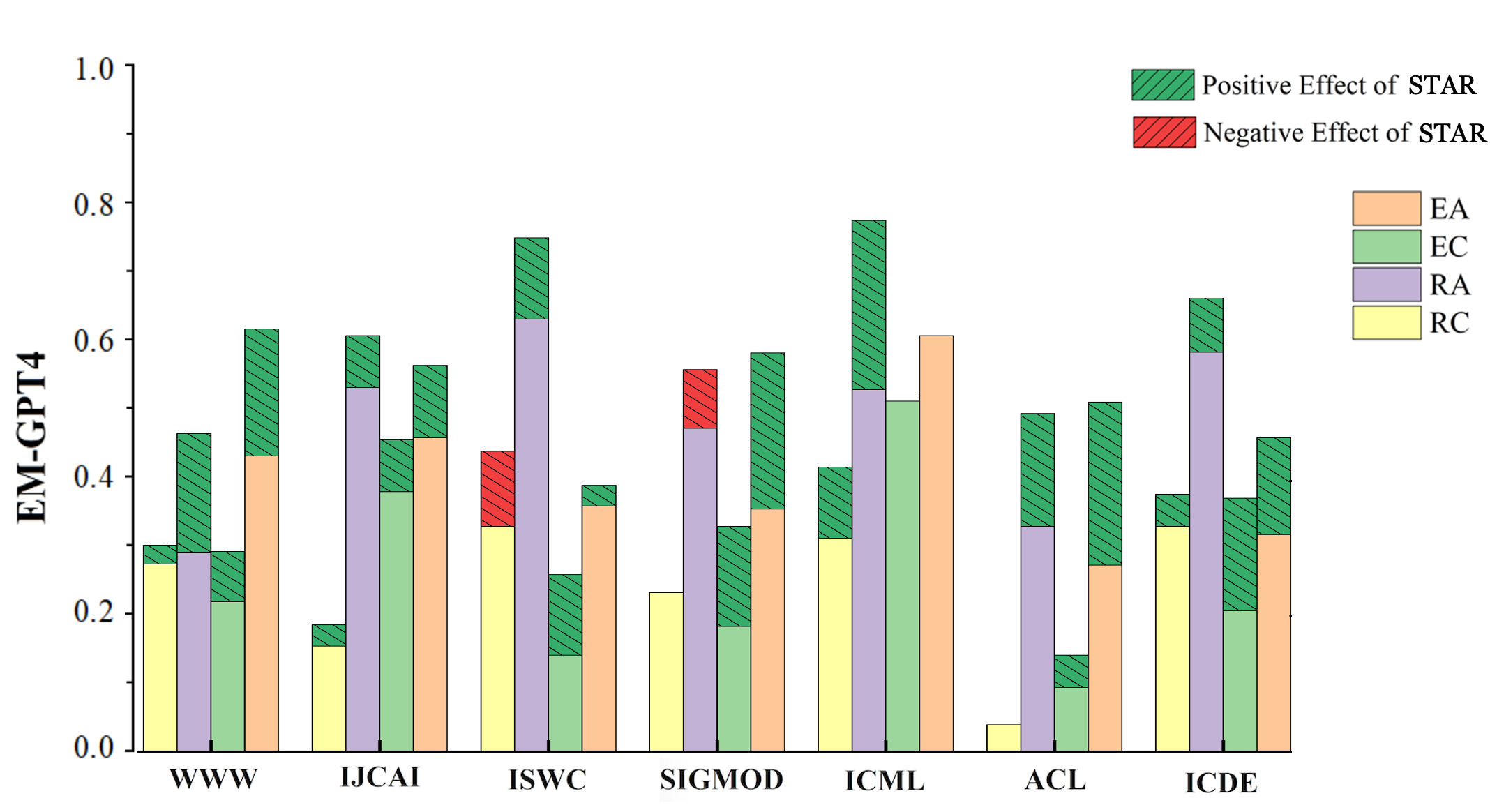}
  \caption{EM-GPT4 metrics across conferences using text-embedding-ada-2 for retrieval and GPT-3.5-turbo for generation. Bars show path retrieval performance; green segments indicate STAR method improvements, red segments show declines.}
\end{figure}

\textbf{Effect of Different Conference} Fig.~\ref{fig:conferences} shows the performance across various conferences using text-embedding-ada-002 and gpt-3.5-turbo as the retriever and generator, respectively. Key observations include: (1) There is notable variability in question difficulty across conferences, highlighting the diversity of our dataset. (2) Significant differences in difficulty are apparent between conferences; for example, the average EM-GPT4 score at ICML is 94.9\% higher than at ACL,
underscoring the importance of accounting for conference-specific characteristics in question-answering research. 
(3) Except for \textit{reasoning-atomic} questions at SIGMOD and \textit{reasoning-complex} questions at ISWC, our STAR method consistently outperforms traditional path retrieval, demonstrating its versatility and effectiveness across different conferences and question types.


\vspace{-2mm}
\section{Related Work} In academic data science, foundational resources such as CiteSeerX\cite{giles1998citeseer}, a digital library for scientific literature, and Unarxive\cite{saier2020unarxive}, which hosts over a million documents from arXiv.org, are crucial for scholarly communication. Zhang et al.\cite{zhang2023effect} developed Maple, a benchmark for tagging scientific literature across 19 disciplines. However, there remains a notable gap in benchmarks specifically designed for academic conference QA, despite the increasing diversity and volume of literature datasets.

Simultaneously, augmenting language models with data from various knowledge bases has significantly improved performance across many NLP tasks\cite{guu2020retrieval, lewis2020retrieval}. Techniques such as Atlas\cite{izacard2022atlas}, which fine-tunes an encoder-decoder model with a retriever, and RETRO\cite{borgeaud2022improving}, which integrates retrieved texts into a decoder-only model, utilize large volumes of unstructured text. Other approaches like REPLUG\cite{shi2023replug} and FLARE\cite{jiang2023active} dynamically retrieve information based on context, treating LLMs as black boxes. In structured knowledge, methods include extracting triples from knowledge graphs for KGQA tasks\cite{sen2023knowledge, hu2023chatdb} and converting them into textual prompts for LLMs\cite{wu2023retrieve} 
However, the use of hierarchical data such as tree-structured data in retrieval augmentation is still limited. 
\vspace{-2mm}
\section{Conclusion}
In this work, we developed the ConferenceQA dataset, which organizes recent academic conference information into a tree-structured format to support question answering. We introduce a novel approach, STAR, that enhances question-answering performance by generating textual descriptions for each path within the tree, effectively utilizing both structural and textual data. The ConferenceQA dataset and STAR method have advanced the development of robust and adaptable academic conference question-answering systems. Future efforts will focus on integrating LLMs with tree-structured data to improve domain-specific knowledge access and reasoning.
\vspace{-2mm}
\section*{Acknowledgements}
This work is founded by National Natural Science Foundation of China (NSFC62\\306276), Zhejiang Provincial Natural Science Foundation of China (No. LQ23F02\\0017), Yongjiang Talent Introduction Programme (2022A-238-G),  Ningbo Natural Science Foundation (2023J291), and Fundamental Research Funds for the Central Universities (226-2023-00138). 


%
%
%
%

\end{document}